# Metastatic Breast Cancer Prognostication Through Multimodal Integration of Dimensionality Reduction Algorithms and Classification Algorithms

Bliss Singhal[1], Fnu Pooja[2]
[1]Bellevue College, Washington, [2]Google, Washington, +poojasn@google.com

## Abstract

Machine learning (ML) is a branch of Artificial Intelligence (AI) where computers analyze data and find patterns in the data. The study focuses on the detection of metastatic cancer using ML. Metastatic cancer is the point where the cancer has spread to other parts of the body and is the cause of approximately 90% of cancer related deaths. Normally, pathologists spend hours each day to manually classify whether tumors are benign or malignant. This tedious task contributes to mislabeling metastasis being over 60% of time and emphasizes the importance to be aware of human error, and other inefficiencies. ML is a good candidate to improve the correct identification of metastatic cancer saving thousands of lives and can also improve the speed and efficiency of the process thereby taking less resources and time. So far, deep learning methodology of AI has been used in the research to detect cancer. This study is a novel approach to determine the potential of using preprocessing algorithms combined with classification algorithms in detecting metastatic cancer. The study used two preprocessing algorithms: principal component analysis (PCA) and the genetic algorithm to reduce the dimensionality of the dataset, and then used three classification algorithms: logistic regression, decision tree classifier, and k-nearest neighbors to detect metastatic cancer in the pathology scans. The highest accuracy of 71.14% was produced by the ML pipeline comprising of PCA, the genetic algorithm, and the k-nearest neighbors algorithm, suggesting that preprocessing and classification algorithms have great potential for detecting metastatic cancer.

## 1. Introduction

Cancer is a devastating global disease. It is one of the leading causes of death today, and around 40% of people will be diagnosed with it in their life. Metastatic cancer is the point where the cancer has spread to other parts of the body and is the cause of 90% of all cancer related deaths [1]. This type of cancer most commonly spreads through the lymphatic system, which protects the body from outside pathogens. The cancer cells travel with the lymph fluid and forms new tumors in the lymph nodes as well as other parts of the body [2]. Metastatic cancer is quite difficult to detect in this method. Doctors must examine hundreds of minuscule images derived from the pathology scans daily, leading to them missing metastases as much as 60% of the time. Therefore, metastatic cancer detection needs a new method. One would think that technology would be utilized to its full potential to combat cancer and improve care to its patients, but this has not been the case. Despite the potential of Machine Learning (ML) to detect cancer as well as other diseases, current cancer detection methods still only depend on the evaluation of a human pathologist, resulting in inefficiency and inaccuracy [3].

Using Artificial Intelligence (AI) to detect cancer has recently been gaining much popularity and there have been numerous studies that prove that AI has the ability to detect cancer more accurately than humans [4, 5, 6]. For example, there was 2016 research conducted by the Harvard Medical School and MIT students that used deep learning and neural networks to detect metastatic cancer [6]. Additionally, Google AI developed a ML algorithm in 2018 that also used deep learning to detect metastatic breast cancer [7]. Other methods, such as different preprocessing and classification algorithms, although quite popular in other fields of research, have not been given sufficient attention. Because of this, this study was conducted to evaluate the potential of these methods to detect cancer. This study proposes a novel approach to analyze the ability of preprocessing, evolutionary, and classification algorithms for detecting metastatic cancer. Specifically, my approach uses principal component analysis, genetic algorithm, and logistic regression to detect metastatic cancer with a higher accuracy than that of a human pathologist.

## 2. Dataset

This study used the PatchCamelyon (PCam) dataset which was derived from the Camelyon16 dataset. This dataset was compiled by the International Symposium of Biomedical Engineering. This dataset was comprised of 32px x 32px pathology images of sentinel lymph node sections. The training set had 220,000 images, and the testing dataset had 57,500 images. Along with the images, the dataset also had a separate CSV file for the training data labels, with one column for the image names, which was a unique id, and the other column for the corresponding label. The label was either a one or a zero, a one indicating at the image has at least one pixel of a cancerous tumor, and a zero indicating that there was not a single pixel of a cancerous tumor.

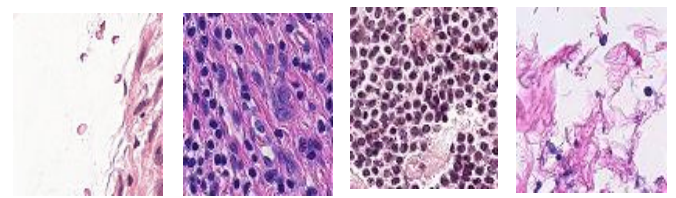

*Figure 1: Pathology image of sentinel lymph node sections from the dataset.*

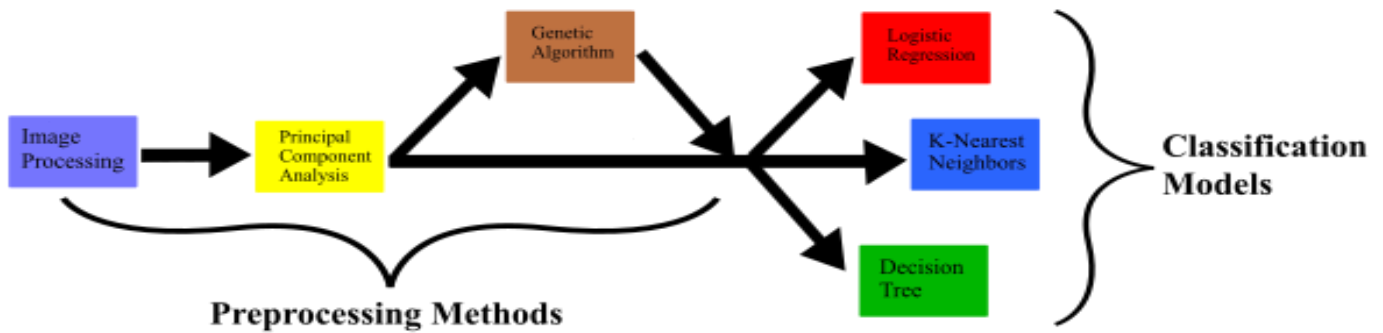
*Figure 2: Machine Learning Pipeline with preprocessing algorithms, and classification algorithms.*

## 3. Methods

Machine learning (ML) techniques were used to classify metastatic cancer in this study. ML is the subsection of AI that derives patterns in the given data and applies them based on different algorithms. There are many different fields in ML, a few main ones are supervised, unsupervised and reinforcement learning. In supervised learning, the data is split into two sections: training and testing data. The training data has labels, indicating the outcome of a particular case. The supervised learning algorithm finds a pattern that associates the training data to its given outcome and uses that relationship to predict the outcomes of the testing data. In unsupervised learning, the data is unlabeled, and the unsupervised learning algorithm finds any patterns it can in the given data. In reinforcement learning, the algorithm is put into an environment which it must adapt to over time. It learns through trial and error. When the algorithm commits an error in the environment, its subsequent consequence, such as a lower accuracy, serve to reinforce the algorithm to avoid making that mistake when the algorithm goes through the environment again. When the algorithm is rewarded, such as getting a higher accuracy, it learns to repeat that action when going through the environment again. In this study, preprocessing of the data was done by using dimensionality reduction techniques. The dimensionality reduction seeks a lower-dimensional representation of input dataset that preserves the salient relationships in the dataset [8]. It helps in compressing the dataset which reduces the storage space requirements and reduces the computational time. Also, it becomes easier to visualize the dataset when reduced to low dimensions. The dimensionality reduction can be done in two main ways: a) feature selection which is done by keeping the most relevant variables from the original dataset, and b) dimensionality reduction where a smaller set of new variables are formed from the original set of input variables by exploiting the redundancy of the input data. There are many different dimensionality reduction algorithms, in this paper, I am proposing the preprocessing of datasets by using Principal Component Analysis, and Genetic algorithm. After the preprocessing, the three classifications algorithms are used to classify the data were k-nearest neighbors, logistic regression, and the decision tree classifier as shown in figure 2.

### 3.1 Preprocessing Methods

Preprocessing methods are used to refine the dataset so that it can be more efficiently used when fed into the classification algorithm.

### 3.1.1 Converting Images to Vectors

The original dataset was in images, which cannot be analyzed directly. The computer can only understand numerical data; hence, the dataset must be converted into numerical data that accurately represents the image. To do this, the images were flattened into a multidimensional array. The first dimension in this array separated the images from each other. The other dimensions represented the rows of that image, and the final dimension represented the RGB color values of a single pixel in that row. Each number in the final dimension in the array represented either an R, G, or B value of one pixel. This process of converting the image into numerical data maintains all the information in that image since it maintains the color of every pixel in the image at its exact location in the image. This means that this data could be used to recreate the original images. This dataset had over 27,000 components for each image, meaning that its dimensionality needed to be reduced.

### 3.1.2 Principal Component Analysis

Principal Component Analysis was used to reduce the dimensionality of the dataset. Principal Component Analysis (PCA) is an unsupervised Machine Learning (ML) algorithm, meaning that it derives patterns from the data without looking for a specific relationship like in supervised learning algorithms. PCA improves the data's readability, reduces the memory needed to store the data, improves the speed, and overall increases the efficiency and usability of the dataset. It reduces the dimensionality effectively by prioritizing dimensions with the highest variability and removing the dimensions with the lowest variability. This dataset had over 27,000 components for each image, far too many to use for ML, so the principal component analysis was necessary to use this dataset. With principal component analysis, these components can be reduced to a much more reasonable number, such as simply 100 components, without reducing the necessary information needed for ML algorithms to predict accurately. The components in the dataset are standardized, meaning that each component's range of values is modified so that they are all equal [9]. This is done so that each component has an equal contribution at the beginning of the analysis. For instance, if a certain component had a range of values leading from 0 to 100, and another component had a range of values only from 0 to 1, the former will be prioritized due to its larger range, introducing bias and inaccuracy in the transformed dataset. The equation used for standardizing a component is $z = $ (value-mean)/standard deviation [9]. Once the data is standardized, Principal Component Analysis algorithm performs covariance matrix contribution to understand the correlation of the data, specifically how the components deviate from the mean and each other. The algorithm transforms the data into a p x p matrix that represents all the possible covariance pairs given the initial components, with p representing the number of dimensions in the data as shown in the below figure.

| Var (x, x) | Var (y, y) | Var (z, z) |
| Cov (x, y) | Cov (y, x) | Cov (z, x) |
| Cov (x, z) | Cov (y, z) | Cov (z, y) |

*Table 1: Covariance Matrix with initial components of x, y, and z*

The covariances show the correlation that the two components have. This is found by first measuring how much each data value strays from the mean of both components. These differences are all added up and divided by the number of data values to get a singular average of the data. If this value is less than the mean of one component, but greater than the mean of another component then the two components have a negative correlation as when one component decreases, the other increases. Else, if the singular average is either less than or greater than the mean of both components, then it has a positive correlation since both components increase and decrease together [9]. When a component is being analyzed against itself then the variance is being calculated as opposed to the covariance. This is the reason that the function for (x, x), (y, y), and (z, z) in the above figure 3 is the variance rather than the covariance. The covariance matrix developed by the algorithm tells how the components are correlated with each other. The data then uses the covariances to compute the eigenvectors and eigenvalues of the components. The eigenvectors and the eigenvalues are used to find the principal components of the dataset.

The principal components are the new components of the data that have as much information compressed into each singular component as possible. They are uncorrelated combinations of the initial components. The number of principal components will be equal to the number of initial components. However, the first principal component will have the most variance, meaning that it will contain the most information, and the second component will have the second most information, and so on. As can be seen in Figure 3, the first principal component has 40% of the total data, while the second component has an approximate of 18% of the data, and the third component has approximately 13% of the data. The percentage of the data that each subsequent component contains decreases.

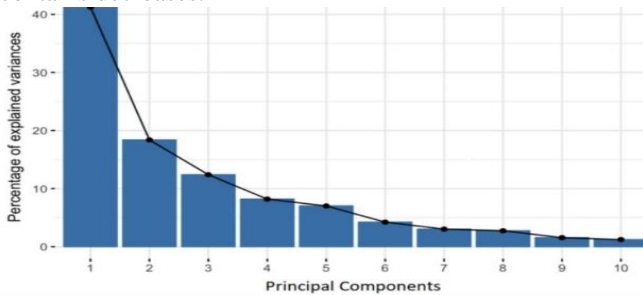

*Figure 3: Variance in Principal Components: graph showing the amount of data stored in each component.*

The principal component themselves do not contain the data.

Instead, they are the eigenvectors of the data and convey the direction of the data with the highest amount of variability. Each principal component is paired with an eigenvalue, to detail the magnitude of variance in the data. The eigenvalues are used to order the principal components in range of importance so that the components with the highest variability are put first. To put the idea of principal components in a visual perspective, the principal components are the new axis for which to analyze the data.

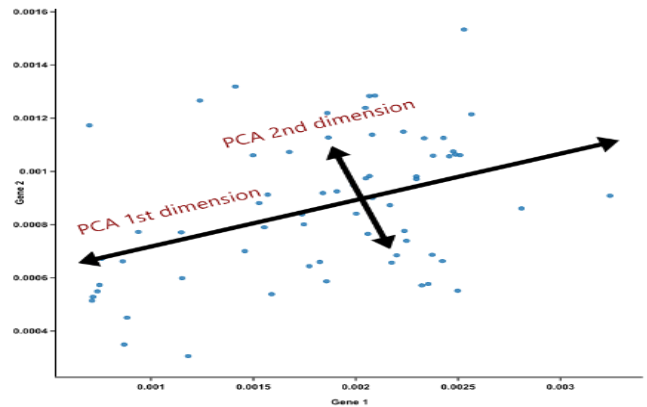

*Figure 4: Visual representation of the process of principal component analysis.*

In this study, PCA analysis was conducted on different sizes of the PCam dataset, starting from a size of 50000 images, and going to a size of only 195 images, with each subsequent dataset size being half of the size before it. As shown in the figure below, the first eight principal components were projected into images.

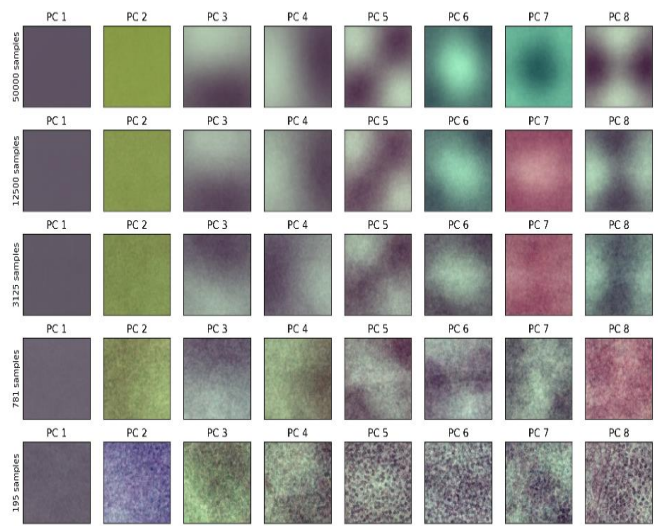

*Figure 5: The first eight principal component projections with different dataset sizes.*

As shown in the figure above, the first and most important component is simply a single color, gray. The second most important component showed only the color green. These two colors, green and gray, had the most variance in the dataset. The third, fourth, and fifth principal components showed different directions of a gradient going from the gray into light green. These gradients were the next most important components in the dataset. Principal components six and seven showed a radial gradient, and the last principal component showed a hyperbolic shape. It is important to note that the principal components were more detailed and specific as for smaller dataset sizes, and the principal components were more universal and generic with larger dataset sizes. With more generic principal components, its projections onto the actual dataset would yeild more accurate results, for it is less specific and can be applied to a larger number of images. This could be one reason why having larger dataset sizes has proved to result in overall higher accuracies in the ML algorithm, as shown in the figure below.

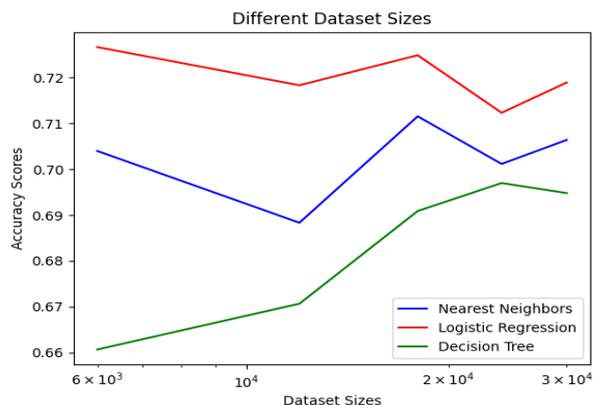

*Figure 6: Accuracy scores for the three classifiers with different dataset sizes.*

Generally, the three lines, representing the three classifiers, rise as the dataset sizes increase, meaning that larger dataset sizes resulted in greater accuracy for in this study. The only parameter for PCA was the number of principal components to select. To find the optimal value for this parameter, the accuracy of three classification algorithms was found with the projections of the different numbers of principal components fed into it as training data.

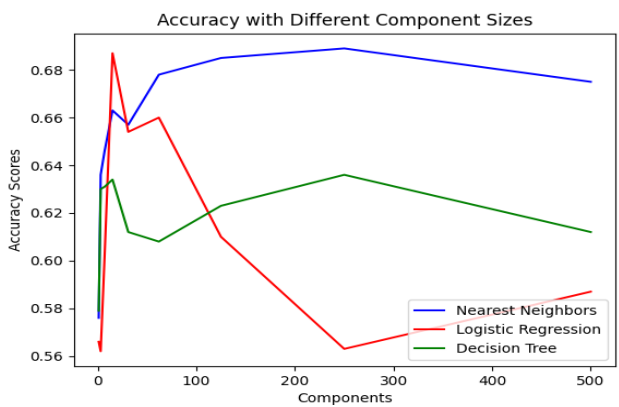

*Figure 7: Accuracy with different number of principal components.*

The maximum number of principal components tested was 500 principal components, with each subsequent number of components being half of the component size of the one prior to it. The dataset size that the PCA algorithm was trained on for generating these results was 600 images. As shown in the figure above, the decision tree classifier and K-nearest neighbors followed a similar trend, where it first had dramatic drops and rises, but after approximately 70 principal components, its accuracies had a gradual rise and after reaching 250 components, its accuracies had a gradual decline as well. The highest accuracy was around 70% at 250 principal components with the K-nearest neighbors classification algorithm. The reason that higher numbers of principal components had lower accuracies is because of overfitting. When the classification algorithm is given too much data to train on, the algorithm overfits, meaning that it memorizes the training data and specifically the noise in the data, meaning that it is unable to adapt to the testing data given, leading to lower accuracy. Another issue of having too many principal components is that it uses too much memory and processing power when fed into the classification ML algorithm, making the process take much more time. In contrast, the component numbers that were less than 250 components had subsequently lower and lower accuracies because the classification algorithm was not given enough data to train on. Because of this, the algorithm was not able to establish the best relationship between the training data and the training data labels, so it gave a lower accuracy. The optimal number of components for principal component analysis in the study was found to be 250 principal components.

It is also interesting to note that the k-nearest neighbors algorithm had the highest accuracy and logistic regression had the lowest accuracy. This was unexpected, considering that logistic regression usually had the highest accuracy when testing all the other parameters in the algorithm.

### 3.1.2 Genetic Algorithm

The second preprocessing method that was used was the genetic algorithm. The genetic algorithm is a stochastic search algorithm that is designed to search for the optimal solution given a set of possible solutions and certain constraints. Genetic algorithms fall under the category of evolutionary algorithms, which are search algorithms based on Darwin's theory of evolution and are modeled after natural selection, otherwise known as "survival of the fittest." These types of algorithms have gained much popularity due to their speed, flexibility, and efficiency in solving complex optimization problems.

The genetic algorithm models evolution that occurs in the real world, looking specifically at the genome. It simulates the genetic changes in reproduction as well as genetic mutations. It creates multiple generations with each generation being more optimal than the one before, due to the theory of natural selection, that the fitter solutions will survive and thrive [10]. First, the genetic algorithm generates several initial possible solutions, generated randomly based on the components given to the algorithm. Each possible solution is called a chromosome and this initial set is called the initial population or Generation 0. The chromosome has a binary encoding that represents the possible solution. It is an array of genes, with a gene being either a random one or a random zero. The one indicates that the corresponding component in the array of components given to the algorithm is used in the solution, while the zero indicates that the particular component is unused in the solution. In this study, the components were the principal components generated by principal component analysis. The gene in the chromosome represents one component, and if that gene had a value of one, then that component is still being included in the dataset, and if that gene was a zero, then the component is excluded and taken out from the dataset. The length of the chromosome, therefore, matches the number of components given. The probability of the gene being a one or a zero was at 50% in the study, but this value can be modified. Once the initial population is created, it is evaluated by the fitness function given by the user. The fitness function is used to evaluate the fitness of a certain solution or how well the solution solves the problem. In this case, the fitness function was the accuracy when the included components in the chromosome was fed into the logistic regression classification algorithm. This is because a higher accuracy indicates that the combination of principal components was had a greater ability to classify cancer. The generation, which is an array of the chromosomes, is then sorted in descending order based on its fitness score, or accuracy. The first two chromosomes indicate the fittest two solutions, so they are added to the new generation since the algorithm does not want to lose its best solutions. The algorithm then performs the single point crossover function, which simulates the creation of offspring from parents. It selects only half of the chromosomes in the population to be parents, with the fitness score being the weights for this selection, to simulate how the fitter solutions are much more likely to survive and have offspring, while the less fit solutions die off before they can do so. For each pair of chromosomes that are parents, the algorithm generates a random index to split the array of genes that both chromosomes consist of. Once each parent chromosome is split into two arrays, the second array of one parent's chromosome is swapped with the other parent, creating two new chromosomes. These new chromosomes are called the children. The children's chromosomes are a combination of their parent's chromosomes, replicating the process of evolution. The children's chromosomes usually have a higher fitness than their parents. The children's chromosomes are not fully developed yet, for the algorithm then simulates genetic mutation for the children. This is to further diversify the chromosomes and differentiate them from their parents by adding new genetic data. Since genetic mutation does not occur for every child, there is a certain genetic mutation probability that the user inputs. This probability defines how likely a genetic mutation is to occur for a child's chromosome. In this study the mutation probability was 25%, because the general mutation probability is set between 20% and 30%. If a genetic mutation is set to occur, the algorithm randomly selects one gene in the chromosome and flips the value. If that value was originally one, that value is now zero, and if that value was originally zero, that value is now one.

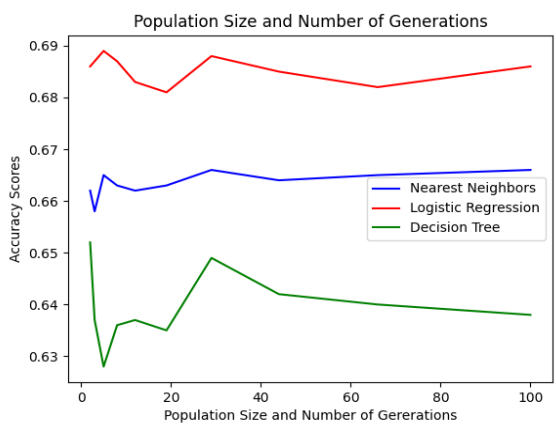

*Figure 8: Accuracies with different population sizes and different number of generations.*

This accurately simulates genetic mutation as shown in figure 8 because it occurs randomly in the chromosome and makes a change that is significant but not dramatic, for genetic mutations in the real world tend to be slight as well. After all the necessary mutations occur, the children's chromosomes are complete and added into the new generation. This process of reordering the generation, keeping the two fittest chromosomes, creating the children, and simulating genetic mutation, and adding the children to the new generation repeats until either the maximum number of generations, which is set by the user, is reached, or it finds the most optimal solution possible, which in this case, was to have a fitness accuracy of 100% [10].

There were many parameters in the genetic algorithm: the population size, the maximum number of generations, the fitness function, the probability of the gene becoming a one or a zero when generating the initial population, and the mutation probability. To find the optimal population size and the maximum number of generations, the genetic algorithm was repeatedly tested with different numbers for population sizes and number of generations. These two variables were tested at the same time because changing one would influence the other, so finding the optimal value parameter separately for each parameter would not truly be its optimal value. The other parameters, such as the dataset size, the number of initial components, fitness function, mutation probability, and the probability of a gene being a one or a zero were all constant, with their values respectively being 600 images, 15 initial components, logistic regression as the fitness function, the mutation probability of 25%, and the probability of a gene value being one or a zero as 50%.

The components selected by the genetic algorithm were fed into the three classification algorithms to get its accuracy. The slopes of the three graphs did not vary greatly, meaning that the population size and the number of generations did not affect the accuracy of the classification algorithms as much. As per these results, the optimal population size and number of components is 29. In addition, when isolating these two parameters, logistic regression overall has a much higher accuracy than k-nearest neighbors and decision tree, while before, when analyzing only different number of principal components, K-nearest neighbors had the highest accuracy throughout. This may be because logistic regression was used as the fitness function, so the genetic algorithm adapted its chromosomes to support logistic regression, and these changes led to more inaccuracy in the other classification algorithms. The genetic algorithm was implemented to test whether having a second preprocessing method yields better results. The results were compared with the classification algorithms without implementing the genetic algorithm, and with implementing the genetic algorithm. The parameters were all kept constant for both cases with same values. This was calculated three times because the genetic algorithm would give different results due to the randomness in the algorithm.

In the first graph as shown in the below figure 9, there was not any difference in accuracy between implementing the genetic algorithm or not, except only a very slight increase in accuracy for the k-nearest neighbors algorithm. However, in the second graph in below figure, the genetic algorithm had a significant improvement in the accuracy of the decision tree classifier, and a slight improvement in the accuracy of k-nearest neighbors, which had the highest accuracy scores out of the three classification algorithms.

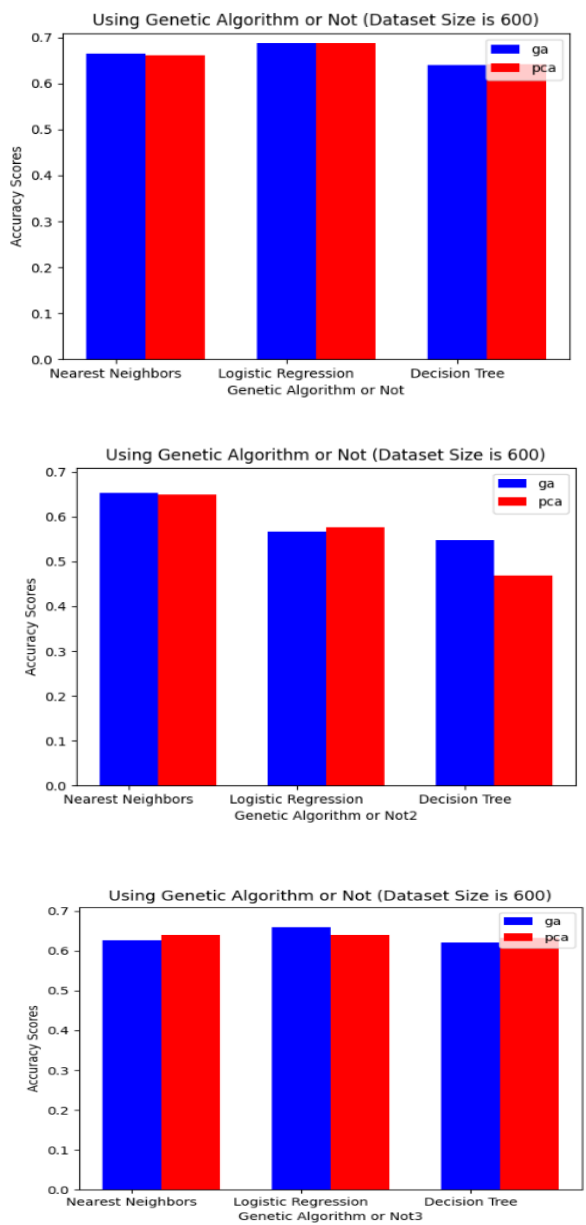

*Figure 9: The accuracy score of implementing genetic algorithm or not for all three classifiers is shown.*

The accuracy without the implementation of the genetic algorithm was slightly greater than the accuracy with the implementation of the genetic algorithm for logistic regression. Lastly, in the third graph as shown in the above figure, the implementation of genetic algorithms improved the accuracy of logistic regression and generated overall the highest accuracy out of all three classification algorithms. In addition, there were many other drawbacks when using the genetic algorithm. First, there were too many parameters that could affect its result and finding the optimal value for each is quite time consuming and requires a fair amount of testing. Furthermore, each test uses quite a lot of memory and processing power, and the program is very slow to execute. However, implementing the genetic algorithm yielded higher accuracies for all the three classification algorithms, so it was deemed necessary to include it in the ML pipeline.

### 3.2 Classification Methods

ML classification algorithms are used to classify data into categories. This is done using supervised learning and finding the relationship of the training data to the categories, given by its labels. This relationship model is then used to classify the testing data.

### 3.2.1 K-Nearest neighbors

The k-nearest neighbors algorithm is one of the most widely used classification algorithm due to its simplicity and efficiency. It classifies the testing data based on its similarity to the training data. It is considered a lazy learner because there is no training phase necessary, for it straightaway compares the testing data to the training data without needing to learn from the training data itself. Even though it is more suited to working with continuous data, in other words, as a regression algorithm, it still can be used for binary classification.

The k-nearest neighbors algorithm examines each testing case and compares each of the component's data values to that of the training cases. It calculates the similarity for each one using the distance formula for finding the distance between the two values. It most commonly uses the Euclidean distance formula, which is $d(p, q) = \sqrt[2]{\sum_{i=1}^{k}(p_i - q_i)^2}$. p and q represent the two cases: the training case and the testing case, and k represents the number of components [11]. The algorithm must calculate the similarity for each component to find the nearest neighbors to the testing case. This process repeats for all the training cases in the training dataset. Once the similarities are calculated, the k-nearest neighbors algorithm finds the number of training cases with the smallest similarities, since this indicates that their component values are the closest together. The number of training cases it determines is based on the number of nearest neighbors that the user inputs into the algorithm. The algorithm then classifies the testing case based on the majority of the nearest neighbors' classifications [11]. For example, if the testing case has three nearest neighbors, and two of them are classified as having cancer, while the third is classified as not having cancer, the testing case will be classified as having cancer as well. If there is an even split of the classifications of the nearest neighbors, say two nearest neighbors are classified as not having cancer and two nearest neighbors are classified as having cancer, the algorithm will eliminate the farthest nearest neighbors until a majority is found. The k-nearest neighbors algorithm had only one parameter, the number of nearest neighbors the algorithm must use to classify the testing data. To find the optimal parameter, the same technique was used as before, of repeatedly testing with different values for the number of nearest neighbors with the optimal value being the one with the highest accuracy. For finding the interval for within which to test the parameter values, the square root of the number of training cases was used, since that has been reported to give the optimal number of nearest neighbors. In this study, with 220,000 training cases, the number happened to be 160, thus, the interval was (1, 320), which each subsequent value being the double of the value before. However, it was noticed that the number of nearest neighbors had no effect on the accuracy score, as shown in the figure below.

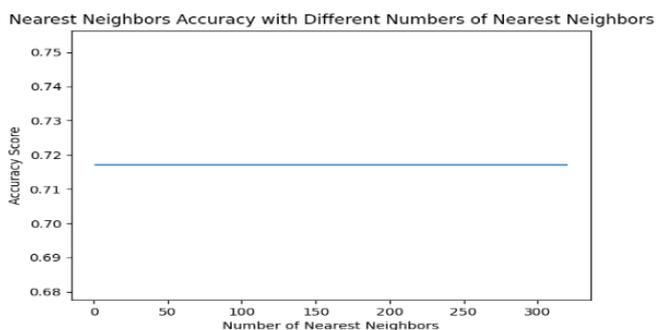

*Figure 10: Accuracy score of the k-nearest neighbors classifier with different number of nearest neighbors.*

As shown in previous figures (Figure 7 and 9), the k-nearest neighbors algorithm had an equal performance to logistic regression and performed significantly better than the decision tree classifier. This result was unanticipated since the k-nearest neighbors algorithm was less tailored to binary classification than the other algorithms, so it was expected that the other algorithms would yield higher accuracy.

The k-nearest neighbors algorithm has some drawbacks as well: it is not adequate for large training datasets with many components and training cases, since it must compute similarities for each component of each testing case for each training case. Therefore, this process uses a lot of processing power, memory, and time. It also uses a single parameter for the number of nearest neighbors. This is an unidentified variable which can greatly affect the nearest neighbors algorithm's accuracy and requires much additional testing to find its suitable value. All in all, the k-nearest neighbors algorithm was important to test in the study and was quite accurate, but was not the most suitable classification method for classifying cancer.

### 3.2.2 Logistic Regression

Similar to the k-nearest neighbors algorithm, logistic regression is a simple and efficient algorithm, but is designed primarily for classification problems, and most specifically towards binary classification. It is a transformation of linear regression, meaning that it has the methodology of linear regression but tailored towards classification problems [12].

Since logistic regression is a form of linear regression, it tries to find a linear relationship between the components, or independent variables in the training data and the output value, or label, which is considered the y-value. Logistic regression uses a sigmoid function, $= \frac{1}{1+e^{-x}}$, to find the relationship between the training data components and the labels. To have a function that can be optimized, it uses a transformed version of the sigmoid function: $y = \frac{1}{1+e^{-\theta^\top x}}$ as its logistic function [13]. The algorithm then tries to find the most optimal value of $\theta$ to find the most accurate relationship between the component (x) and the label (y). Since the dataset in this study has multiple components, the sigmoid function would be $y = \frac{1}{1+e^{-(\theta_0+\theta_1 x_1+\theta_2 x_2+\cdots+\theta_n x_n)}}$, with n representing the number of components in the training dataset, or 27,000. The function must find the optimal parameter or the $\theta$ value for each $x_a$, with $a$ representing the index for a single component. In this way, it can account for each component value in finding the most accurate mathematical relationship between the training data and the training data's labels. Therefore, it is able to find the probability of each component in the testing data directly affecting the classification of the label [13]. When the x-values and corresponding y-values are projected onto a coordinate plane, the resulting graph is the shape of an s, with asymptotes at x=0 and x = 1 as shown below.

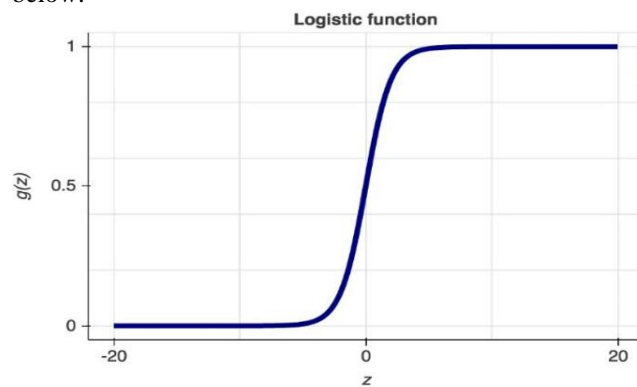

*Figure 11: The graph of the logistic function with the asymptotes at x=0 and x = 1.*

The sigmoid function, with the parameter values optimized, represents a decision boundary for the two categories in binary classification. When the value of y in the sigmoid function is over the threshold of 0.5, it will classify it as one category, but if it is under the threshold, the algorithm will classify it as the other category. In this study, the categories would be having cancer or not having cancer.

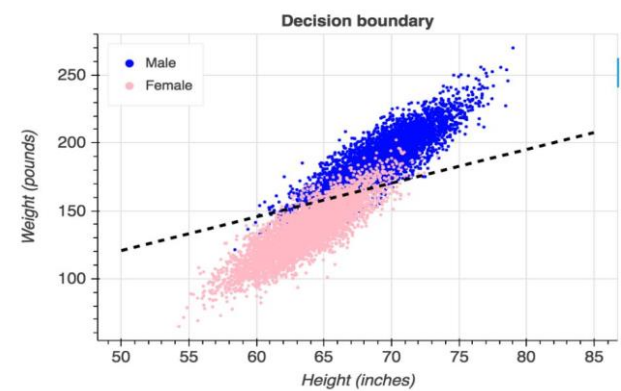

*Figure 12: Example of the sigmoid function being a decision boundary for the data.*

The logistic regression classification algorithm performed the most accurately out of the three algorithms, as shown in the diagrams before, such as the graphs plotting the accuracy scores for implementing the genetic algorithm as well as the graph for the genetic algorithm, which showed the optimization of the number of generations and population size. This was one of the main benefits of implementing the logistic regression. It had one critical parameter: the solver, which was the algorithm used in the optimization problem. The solver was instrumental in the performance of logistic regression. Originally, the parameter was set to the default value, Limited-memory-Broyden–Fletcher–Goldfarb–Shanno

(LBFGS), but it yielded an error because it failed to converge, and the program was quite slow. The solver's value was then changed to saga because of saga's ability to work with large datasets that have high dimensionality, and logistic regression. As a result, logistic regression with saga performed much more efficiently. Another issue with logistic regression is that since it is derived from linear regression, it assumes independent variables have a linear relationship with the dependent variables. This assumption is not always correct and can cause it to be more inaccurate with data that has a different mathematical relationship, such as quadratic or trigonometric. Overall, due to its highest accuracy among the three classification algorithms, as shown in previous figures (Figure 6, 8, and 9), the logistic regression algorithm has the potential to be the most optimal classifier used in this study.

### 3.2.3 Decision Tree Classifier

The decision tree classifier is the final Machine Learning (ML) algorithm implemented in the study. It is primarily a classification algorithm but can also be used for regression. This algorithm was implemented in the study since it is a much more complex and sophisticated algorithm than k-nearest neighbors and logistic regression and it is quite similar to neural networks.

This classifier is entirely based on asking a series of questions to find the result. These questions are organized in a decision tree algorithm to show which questions are asked first and which questions are asked based on the answers to the previous questions. The decision tree is made up of nodes and edges. Nodes indicate the questions that are derived from the components of the dataset, and the edges indicate the answers to the question, and what node comes after or before that particular node. There are three types of nodes in the decision tree: the root node, internal node, and leaf node [14]. The root node indicates the first question that is asked, so there are no incoming edges, since there is no question asked before it and any number of outgoing edges, because it can lead to as many or as less different answers. The internal node is what the decision tree consists mainly of and has one incoming edge and two outgoing edges because there are only two possible answers for the questions. The leaf nodes are the bottom of the decision tree, where there are no more questions to be asked and the training case is classified. It has one incoming edge and no outgoing edges.

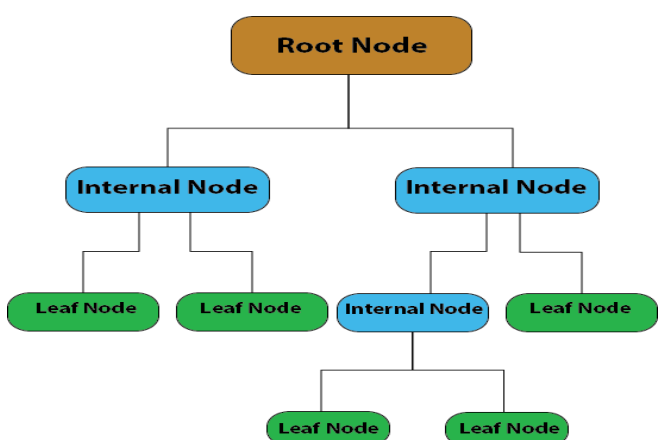

*Figure 13: A diagram showing the structure of the decision tree classifier.*

On a lower level, the decision tree classifier does recursive partitioning, where it keeps splitting the data into smaller and smaller subsets. The root nodes and internal nodes serve as criteria on how to split the data. This criterion is derived with certain components from the dataset. There are different types of splits for the data, with the main ones being the binary split and nominal split. A binary split is when the dataset is split into two subsets based on a condition with only two possible answers. A nominal split is when the data is over two possible answers, so the algorithm does several binary splits to separate the subsets based on each of those possible answers.

To select the best nodes and evaluate how beneficial the split was in classifying the data, there are several different criteria methods that can be used. One method is information gain using child node impurity. The information gain formula is $IG(D_p, f) = I(D_p) - \sum_{j=1}^{m} \frac{N_j}{N_p} I(D_j)$. $f$ is the component used to split the data, $D_p$ is the data subset of the parent nodes, $D_j$ is the data subset of the child node, $N_p$ is the number of samples in the parent node dataset, $N_j$ is the number of samples in the child node dataset, and $I$ is the impurity measure [15]. The impurity of the parent nodes is compared to the impurity of the child nodes, and if the impurity of the child nodes is lesser, than the test condition was beneficial, but if the impurity of the child node is greater than the impurity of the parent nodes, than the test condition should not be implemented because it causes the data to be less skewed. This impurity can be calculated using child node impurity or entropy [15]. The child node impurity method is based on class distribution. The smaller the impurity, the more skewed the distribution is. The goal is to get an impurity of zero, which is a distribution of (0, 1), meaning that the data is completely skewed to one side and separated based on its classification. Entropy is the variance in the data and calculated based on the collection of data points at each node. As with the child node impurity method, the less the entropy value, the more skewed the class distribution is.

The other benefit of a decision tree classifier is that there are no parameters, so there is no additional testing needed to find the optimal parameter. However, this classifier consistently performed the worst out of the three classification algorithms and had the lowest accuracy score. This may be because it has a problem of overfitting the data, and the training data set was quite large, with over 6,000 samples. Because of this, it should not be implemented to classify the data.

## 4. Results

For detecting metastatic cancer, the study tested five different Machine Learning (ML) algorithms: principal component analysis (PCA), genetic algorithm, k-nearest neighbors, logistic regression, and the decision tree classifier. Principal component analysis and the genetic algorithm were both preprocessing methods to reduce the dimensionality of the data, and the result was that both were necessary and improved the accuracy of the classification algorithms. The parameters and their optimal values were the training dataset's size being 6,250 samples, PCA's number of principal components being 250, the genetic algorithm being implemented with its number of generations and generation size being 29, and logistic regression's solver being saga. The resulting pipeline is that the algorithm starts with converting the image data into a numerical array, performing principal component analysis on that array, implementing the genetic algorithm, and finally classifying the testing data with k-nearest neighbors. As for the three classification algorithms, the end result was that with all the parameters optimized, k-nearest neighbors had the highest accuracy of 71.14%, while logistic regression was only a little bit lower at 71%, and the decision tree classifier was dramatically lower, at 64%. This is shown in the below figure.

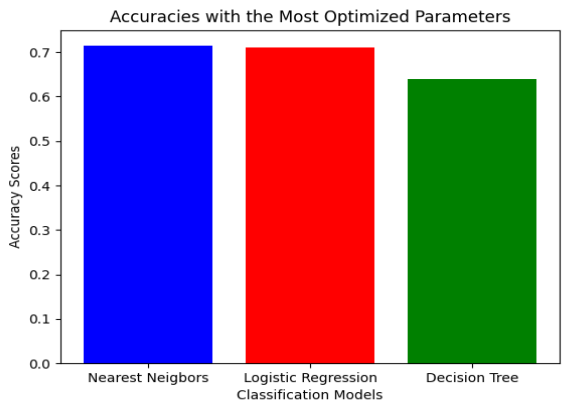

*Figure 14: The graph shows the accuracy scores of k-nearest neighbors, logistic regression, and decision tree classifier algorithms where all three used preprocessing steps with principal component analysis and genetic algorithm.*

Even though logistic regression more often had the highest accuracy when a single parameter was being optimized, k-nearest neighbors had the highest accuracy overall when all the parameters were optimized. The parameters and their optimal values were the training dataset's size being 6,250 samples, PCA's number of principal components being 250, the genetic algorithm being implemented with its number of generations and generation size being 29, and logistic regression's solver being saga. The resulting pipeline is that the algorithm must start with converting the image data into a numerical array, performing principal component analysis on that array, implementing the genetic algorithm, and finally classifying the testing data with k-nearest neighbors. The result that k-nearest neighbors was the most accurate classifier could potentially be because it was a simpler algorithm and generally, do not overfit the data. It also did better than logistic regression because it supports non-linear data, and the relationship between the training data components and the labels would not have been exactly linear, and logistic regression only finds a linear relationship with the data. Since the pathologists' average accuracy of detecting metastatic cancer is 40% using manual detection, the results showed that the principal component analysis algorithm, the genetic algorithm combined with the classification algorithms, have the potential to detect metastatic cancer more accurately than manual methods.

## 5.  Conclusion

This study concludes that Machine Learning (ML) preprocessing methods and classification algorithms have the ability to detect metastatic cancer more accurately than manual methods. The combination of the principal component analysis algorithm and the genetic algorithm improved the overall accuracy when combined with the classification algorithms. The preprocessing methods reduced noise and the dimensionality in the dataset, and the classification algorithms, specifically k-nearest neighbors, were able to find a relationship between the data's components and the labels. This study's benefits were that it analyzed multiple preprocessing methods and classification algorithms and compared them based on their most accurate scores with the optimal parameter values. It also compared the effect of complex ML classifiers, such as the decision tree classifier against the effect of simple ML classifiers, such as logistic regression and k-nearest neighbors. Since the study was conducted with the same set of training data and was not able to get the accuracy above 71%, the future exploration can be done to use different training and testing datasets. Also, neural networks, and deep learning algorithms can be explored to compare against the classification algorithms.